\documentclass{article}
\usepackage{ijcai22}

\usepackage{times}

\usepackage{soul}
\usepackage{url}
\usepackage[hidelinks]{hyperref}
\usepackage[utf8]{inputenc}
\usepackage[small]{caption}
\usepackage{graphicx}
\usepackage{amsmath}
\usepackage{booktabs}
\title{Shift-Robust Node Classification via Graph Adversarial Clustering}
\usepackage{amsmath, amsfonts, amssymb, xspace, color, amsbsy, caption, adjustbox, amsthm}
\usepackage{tabularx}
\usepackage{tikz, graphicx}
\usepackage{enumitem}
\usepackage{multirow}
\usepackage{cleveref}
\usetikzlibrary{positioning}
\usepackage[linesnumbered,ruled,lined]{algorithm2e}
\usepackage[font=small,labelfont=bf]{caption}
\usepackage{subcaption}
\usepackage{esvect}

\usepackage{changepage}
\usepackage{pifont}
\theoremstyle{plain}
\newtheorem{definition}{Definition}[section]

\newtheorem{thm:eg}{Example}

\newcommand{\Ours}{\textsc{SRNC}\xspace}
\newcommand{\xhdr}[1]{\vspace{1.7mm}\noindent{{\bf #1}}}

\newcommand{\ie}{\emph{i.e.}\xspace} 
\newcommand{\eg}{\emph{e.g.}\xspace} 
\newcommand{\wrt}{\emph{w.r.t.}\xspace} 

\newcommand{\Sum}{\sum\limits} 

\DeclareMathOperator*{\argmin}{argmin}

\def \y {\mathbf{y}}

\def \P {\mathbf{P}}

\def \D {\mathcal{D}}

\def \N {\mathcal{N}}

\def \Q {\mathcal{Q}}

\def \S {\mathcal{S}}

\SetKwRepeat{Do}{do}{while}

\begin{document}
\author{Qi Zhu$^1$ \and Chao Zhang$^2$\and Chanyoung Park$^3$\and
Carl Yang$^4$ \and Jiawei Han$^1$ \\
\affiliations
$^1$University of Illinois Urbana-Champaign, $^2$Georgia Institute of Technology,\\
$^3$KAIST,  $^4$Emory University\\
\emails
{$^1$\{qiz3,hanj\}@illinois.edu, $^2$chaozhang@gatech.edu}, \\
{$^3$cy.park@kaist.ac.kr, $^4$j.carlyang@emory.edu}
}
\maketitle

\begin{abstract}
Graph Neural Networks (GNNs) are de facto node classification models in graph structured data. However, existing algorithms assume no distribution shift happens during testing-time, \ie, $\Pr_\text{train}(X,Y) = \Pr_\text{test}(X,Y)$. 
Domain adaption methods are designed to account for distribution shift, yet most of them only encourage similar feature distribution between source and target data. Conditional shift ($\Pr(Y|X)$) on label can still affect such adaption.
In response, we propose
\textbf{S}hift-\textbf{R}obust \textbf{N}ode \textbf{C}lassification (\Ours) featuring an unsupervised cluster GNN, which groups the similar nodes by \textit{graph homophily} on target graph. 
Then a shift-robust classifier is optimized on training graph and adversarial samples from target graph, which are provided by cluster GNN.
We conduct comprehensive experiments on both synthetic and real-world distribution shifts. Under open-set shift,  we see the accuracy is improved at least $3\%$ against the second-best baseline. On large dataset with close-set shift - ogb-arxiv, existing domain adaption algorithms can barely improve the generalization if not worse.
\Ours is still able to mitigate the negative effect ($>$ 2\% absolute improvements) of the shift across different testing-time.


\end{abstract}

\section{Introduction}
Graph
Neural Networks (GNNs)~\cite{kipf2016semi,velivckovic2017graph,hamilton2017inductive} have
achieved enormous success for node classification. Recently, researchers observe sub-optimal performance when training data exhibit distributional shift against testing data~\cite{zhu2021shift}. In real-world graphs, GNNs suffer poor generalization from two major kinds of shift: (1) open-set shift: there are emerging new classes in the graph (\eg, new COVID variant in community transmission); (2) close-set shift: time-augmented test graph has substantially different class distributions (\eg, time-evolving Twitter and Academia graphs).

In machine learning, domain adaption helps model generalize to target domain when there is data shift between source and target domain. 
For open-set shift, Out-of-Distribution (OOD) detection
methods ~\cite{hendrycks2016baseline} calibrate supervised
classifiers to detect samples belonging to other domains (\eg, different datasets) or unseen
classes. OpenWGL~\cite{wu2020openwgl} and PGL~\cite{luo2020progressive} aim for open-set classification on graph, where they both progressive add pseudo labels for ``unseen'' classes. For close-set shift, inferior performance caused by non-IID training data is first discussed~\cite{ma2021subgroup,zhu2021shift} for GNNs. More general close-set shift is observed in time-evolving (dynamic) graph, where training graph can be viewed as a non-IID sub-graph of test graph. To the light of this issue, the most well-known approach is domain invariant representation learning~\cite{Long2015,Long2017,zellinger2017central}, namely, indistinguishable feature representations between source and target domain. However, the limitation of these methods is obvious when class-conditional distributions of input features change ($\Pr_\text{train}(Y|X) \neq \Pr_\text{test}(Y|X)$) - conditional shift~\cite{zhao2019learning}.

\begin{figure}
    \centering
    \includegraphics[width=0.43\textwidth]{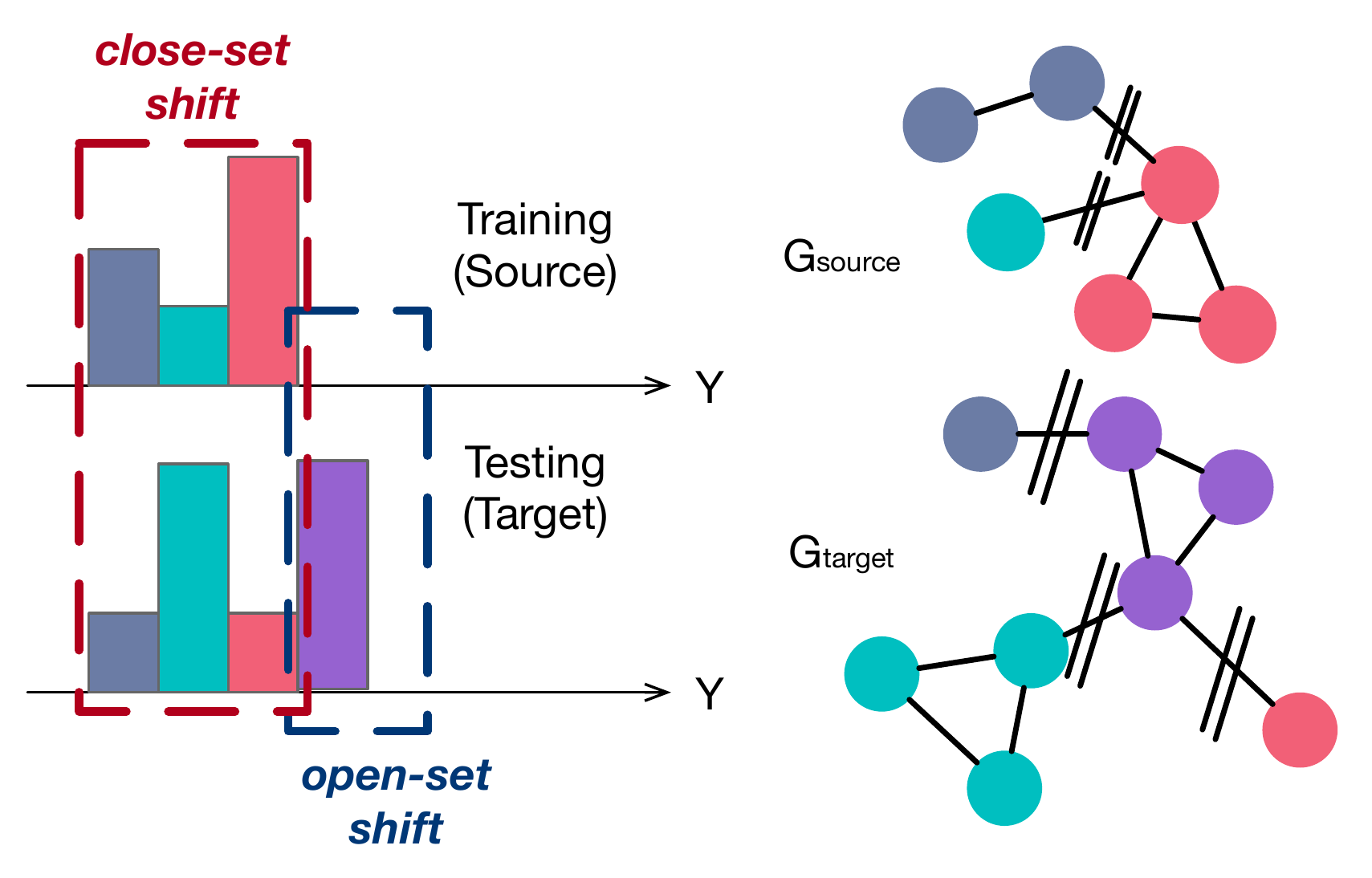}
    \caption{A unified domain adaption perspective for open-set shift and close-set shift. \textit{graph homophily} is universally hold on source and target indicated by $\|$ cut.}
    \label{fig:openset-example}
\end{figure}




Despite the popularity and importance of GNNs,  only a few attempts aim to make
GNNs robust of either open-set shift~\cite{wu2020openwgl,luo2020progressive} or close-set shift~\cite{zhu2021shift}. All previous work do not incorporate graph-specific designs and still suffer from the aforementioned conditional shift (see results of domain-invariant baselines in Table~\ref{tab:close-set-result} for more details). For example, various degree of \textit{graph homophily} (\eg, neighbors with same labels) exhibit in different graphs, whose potential has not been explored against distribution shifts.

In this paper, we propose a unified domain adaption framework for shift-robust node classification, where unlabeled target (testing) graph is used to facilitate model generalization. 
As shown in Figure~\ref{fig:openset-example}, both shifts can be interpreted as one specific distribution shift in the plot.
A major reason behind the ineffectiveness of existing domain adaption techniques for GNN is \textit{the lack of modeling classes in target data}, \ie $\Pr(X^t,Y^t)$. 
Without modeling class distribution on target, it is impossible to mitigate the conditional shift when it happens.
Motivated by \textit{graph homophily}, we propose to use graph clustering to identify latent classes~\cite{bianchi2020spectral,tsitsulin2020graph} by breaking less frequent edges (possibly heterophily) between potential clusters (right side of Figure~\ref{fig:openset-example}). Our proposed model, which we call \Ours (Section~\ref{sec:openssl}), features two graph neural networks (GNNs): an shift-robust
classification GNN$_\Theta$  and an adversarial clustering GNN$_\Phi$
(Section~\ref{sec:model-q}).
In \Ours optimization, the two modules improve each other through adversarial training:
\begin{enumerate}

\item
  \emph{The clustering structure inferred by the
clustering GNN on source data $\Pr_\Phi(C^s|X^s)$
should be close to training conditional distribution $\Pr_\text{train}(Y^s|X^s)$.} For instance, in upper Figure 2, the KL-divergence between above two probabilities are minimized to push the clustering result consistent with training data.

\item
 \emph{In the other direction, the classifier is optimized on both training graph and target graph.} The adversarial samples are sampled from clustering GNN 
 to improve the generalization on target graph. 
\end{enumerate}
As far as we know, it is the first domain adaption framework considers the joint target probability $\Pr(x_i^t,y_i^t)$ on graph.
The joint probability avoids tackling conditional shift on $\Pr(y_i^t|x_i^t)$ and our real-world experiments (see Section~\ref{sec:close-set}) demonstrates that only minimizing feature distribution discrepancy leads to negative transfer.
Furthermore,
convergence of the iterative optimization is theoretically guaranteed in Section \ref{sec:co-train}.

To summarize, we present a domain adaption framework for two kinds of most common data shift in graph structured data. 
Our experiments show that the shift-robust classifier trained in this way can
detect more than 70\% open-set samples on three widely used benchmarks.
For close-set shift, we use early snapshot (prior to 2011) of OGB-Arxiv~\cite{hu2020open} network as training graph and testing on more recent graphs (14-16/16-18/18-20). \Ours can reduce the negative effect of shift by at least ~30\% in testing. We further show progressively utilizing classifier's prediction as pseudo labels (our ablation without clustering component) on target graph~\cite{wu2020openwgl,luo2020progressive} is much worse than proposed method. 

\section{RELATED WORK}

\xhdr{Domain Adaption.}
Domain adaption transfers machine learning models trained on the \emph{source} domain to the related \emph{target} domain. 
The $\mathcal{H}$-divergence \cite{Ben-David2010}  is first developed as a distance function between a model's performance on the source and target domains to describe how similar they are. Then discrepancy measures \cite{mansour2009domain} are invented to measure the generalization bound between source and target domains. To bridge the gap between source and target, Domain Invariant Representation Learning aims to minimize these discrepancy measures on source and (unlabeled) target data by adversarial learning~\cite{ganin2016domain}  or regularization (\eg CMD~\cite{zellinger2017central}, MMD~\cite{Long2015,Long2017}). We also use unlabeled target data in our framework, but we propose a clustering component for adversarial samples specifically for graph structured data.


\xhdr{Open-set Classification.}
Open set recognition and classification~\cite{scheirer2014probability,bendale2016towards} require classifiers to detect \textit{unknown} objects during testing.
A closely related problem is  out-of-distribution
detection~\cite{hendrycks2016baseline}, which focuses on
detecting test samples of rather different distribution than in-distribution (training distribution
by classifier). These studies focus on addressing the overconfidence issue of deep neural networks
for unknown classes, such as OpenMax~\cite{bendale2016towards} and DOC~\cite{shu2017doc}.
Specifically, on graph structured data, OpenWGL~\cite{wu2020openwgl} employs an uncertainty loss in the form of graph reconstruction loss~\cite{kipf2016variational} on unlabeled data. PGL~\cite{luo2020progressive} extends previous domain adaption framework to open-set scenario with graph neural networks.
Yet, none of open-set learning on graph has explored modeling class distribution on target graph by \textit{graph homophily}.

\xhdr{Distribution Shift on GNNs.} There has been a recent surge on discussing GNNs generalization~\cite{garg2020generalization,verma2019stability,ma2021subgroup}. The generalization and stability of GCNs is found to be related with graph filters~\cite{verma2019stability}. Our work mainly focus on the distribution shifts between specific training and testing data. To this end, SRGNN~\cite{zhu2021shift} first adopts two shift-robust techniques - CMD and importance sampling for non-IID training data. However, the previous explorations assume no conditional shift such that adaption is limited or even negative in real-world situations.





\section{Problem Definition \& Preliminary}
\label{sec:problem}
$\mathcal{G} = \{V, A, X\}$ is defined as a graph with nodes $V$, their features ($X \in \mathbb{R}^{|V| \times |F|}$) and edges between nodes (\eg, adjacency matrix $A, A \in \mathbb{R}^{|V| \times |V|}$). 
A $\text{GNN}$ encoder takes node features $X$ and adjacency matrix $A$ as input, aggregates the neighborhood information and outputs representation $h_v^k$ for each node $v$. In its $k$-th layer, for node $i$, the GNN encoder aggregates neighbor information from $k-1$ layer into neighborhood representation $z$:
\begin{equation}
    h_i^k = \textsc{Aggregate} \circ \textsc{Transform}^{(k)} \left( \{ h_j^{k-1}, j \in \mathcal{N}_i\}\right),
\end{equation}
where $h_i^k \in \mathbb{R}^{d}$ is the hidden representation at each layer, $\mathcal{N}_i$ is the neighborhood of node $i$ and $h^0 = X$. 

In graph, \textit{graph homophily}~\cite{zhu2020beyond} indicates densely connected nodes tend to have similar properties.
Recent study has shown \textit{graph homophily} as key factor to the success of graph neural networks on node classification, where nodes with similar representations are likely to have same label. 
\begin{definition} [Distribution Shift in Node Classification]
Given a graph $\mathcal{G}$ and labeled source data $\mathcal{D}^s=\{(x_1, y_1), (x_2, y_2), ...
(x_n, y_n)\}$, a semi-supervised learning classifier computes a cross-entropy loss over $N$ classes,
\begin{equation}
    \mathcal{L} = \frac{1}{|\mathcal{D}^s|} \Sum_{i=1}^n \Sum_{j=1}^{N} y_{ij} \log \Hat{y}_{ij}, \Hat{y}_{ij} = \textbf{softmax}(f(h_i^k))
\end{equation}
Distribution shift occurs if $\Pr_\text{train}(H,Y) \neq \Pr_\text{test}(H,Y)$\footnote{We study the distribution shift in final latent space $H$ and refer $\Pr(H,Y)$ as joint probability in the paper.}.
\end{definition}

Among various possible causes of distribution shift, we are interested in two major shifts in this paper: (1) open-set shift, namely, new classes arise during test-time, \ie $|y_\text{train}| < |y_\text{test}|$. (2) close-set shift, joint probability changes between training and testing, $\Pr_\text{train}(H,Y) \neq \Pr_\text{test}(H,Y)$.




\section{Method}
\label{sec:method}
In this section, we present our framework for shift-robust node classification. 
\Ours contains two adversarial GNN modules (1) a shift-robust classification $\text{GNN}$ $\P_{\theta}$ (Section~\ref{sec:openssl}) and (2)
an unsupervised clustering $\text{GNN}$ $\Q_{\Phi}$ (Section~\ref{sec:model-q}).
The classification GNN is optimized on labeled source graph and pseudo labels (adversarial samples) on target graph for better generalization against possible distributional shifts. Meanwhile, the clustering GNN is optimized on target graph and regularized with training data from source.
Finally in Section~\ref{sec:co-train} and Algorithm~\ref{alg}, we summarize how
we optimize \Ours via adversarial training between both modules. We also provide convergence analysis of the training by Expectation-Maximization (EM).
In the remaining of the paper, we call training data as source and testing data as target interchangeably.

\subsection{Shift-Robust Node Classification}
\label{sec:openssl}

\begin{figure}
    \centering
    \includegraphics[width=0.43\textwidth]{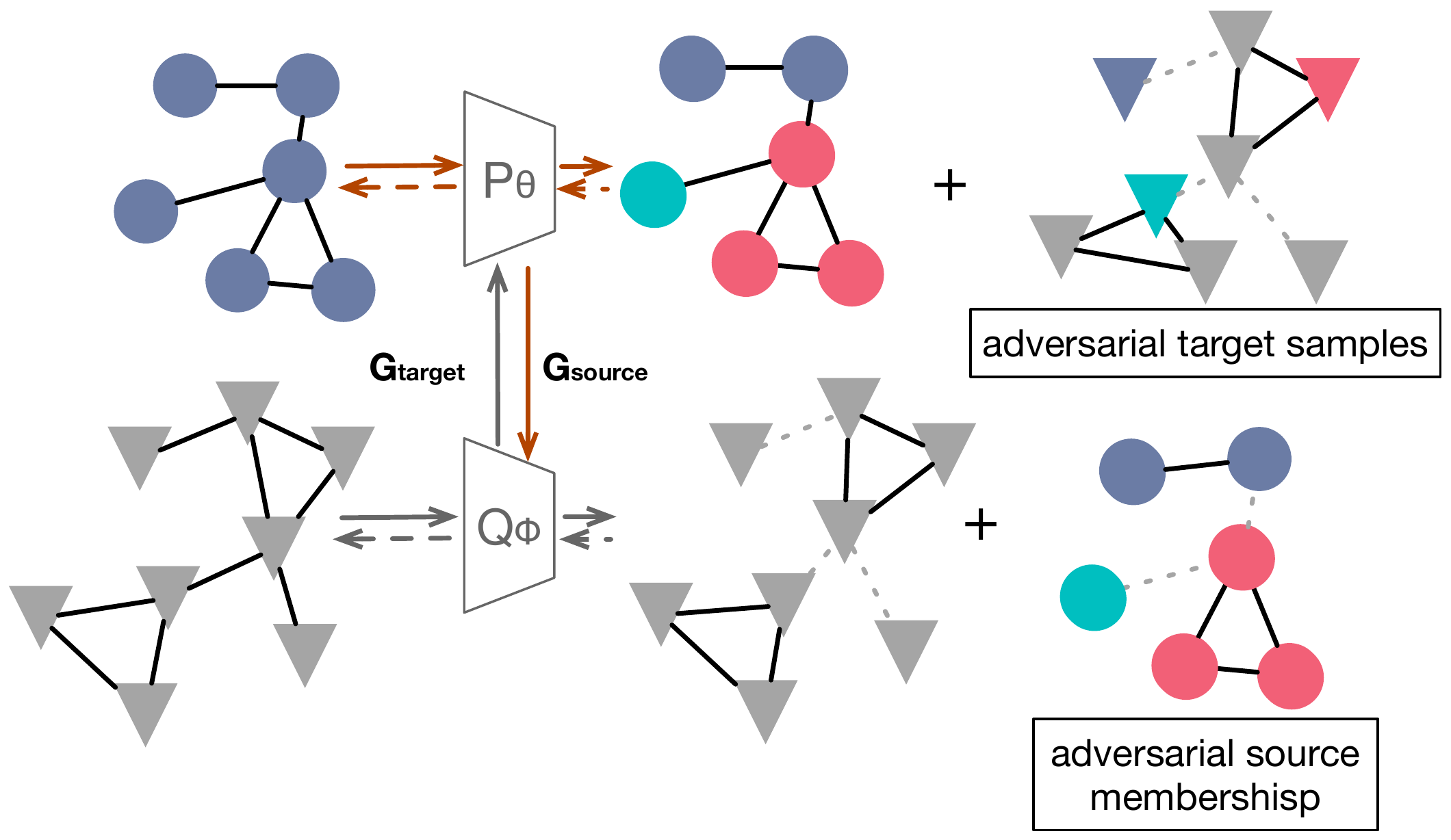}
    \caption{Training framework of \Ours. Solid arrow represents forward pass and dash arrow represents back propagation. The first and second row illustrate the training process of classification and clustering GNNs respectively.}
    \label{fig:framework}
\end{figure}

We first propose the domain adaption loss for shift-robust
semi-supervised classification on distribution shifts.
Classifier $\P_{\theta}$ minimize a negative log-likelihood loss $\mathcal{L}_{\theta,S}$  on training data $(x_i^s, y_i^s)$, and shift-robust loss $\mathcal{L}_{\theta,T}$ on adversarial samples $(x_i^t, y_i^t)$ from target (testing) data.
\begin{equation}
\begin{array}{l}
\label{eq:open-ssl}
    \mathcal{L}_{\theta,S} = \Sum_{i=1}^{|D^s|} - \log \P_{\theta}(y_i^s|x_i^s, \text{A}^s) \\
    \mathcal{L}_{\theta,\Phi,T} = \mathbb{E}_{(x_j^t,y_j^t) \sim \Q_{\Phi}(\cdot |\mathbf{X}^t, \text{A}^t) } \left[ - \log \P_{\theta}(y_j^t|x_j^t, \text{A}^t) \right]
\end{array}
\end{equation}
The sampling process in the second term $(x_j^t,y_j^t) \sim \Q_{\Phi}$ first uniformly sample same amount of nodes as training $\{x_j^t\}_{j=1}^{|\D^s|}$ from target graph. Then we obtain their pseudo labels $\{y_j^t\}_{j=1}^{|\D^s|}$ through cluster GNN $\Q_{\Phi}(y_j |x_j^t, \text{A}^t)$. We will discuss how to align the identity of cluster $\{1..C\}$ and classes $\{1..N\}$ in the beginning of model optimization (Section~\ref{sec:co-train}). In first row of Figure~\ref{fig:framework}, $\P_{\theta}$ is first trained on source graph and jointly optimized on adversarial target samples $\{x_j^t, y_j^t\}$.
Note that source and target graph can be same (open-set shift) or different (close-set shift).

Our framework works for both kinds of distribution shifts that we deem essential in the paper. For open-set shift, we set number of cluster larger than known classes, \ie $C>N$ and $\P_{\theta}$ classify target data into seen $N$ classes and the new $N+1$ class.
In second term of above Equation~\ref{eq:open-ssl}, the samples $y_j^t \in C \setminus N$ from unaligned clusters are mapped into the unified unknown class $N+1$. For close-set shift, we simply set number of cluster and classes equal in two modules.


\subsection{Graph Adversarial Clustering}
\label{sec:model-q}
In this section, we will explain how to approximate the target data distribution $\{x_i^t, y_i^t\} \sim \mathbb{P}^t$ in Equation~\ref{eq:open-ssl} and its connection with \textit{graph homophily} and clustering. 
The sufficient condition of uniform sampling from target $\mathbb{P}^t$ requires an unbiased estimation of $\mathbb{P}^t(y_i^t|x_i^t)$ if we uniformly sample $x_i^t$. Although underlying $\mathbb{P}^t(y_i^t|x_i^t)$ is unknown, similar nodes likely appear in the same cluster based on \textit{graph homophily}. 
In graph clustering, to model node $k$'s
cluster membership $c_k$, we have $ \Q_{\Phi}(c_k|\mathbf{X},\mathbf{A}) \sim
\text{Categorical}(\Phi)$. We thus parameterize $\Q_{\Phi}$ using another GNN's outputs:
\begin{equation}
\label{eq:cluster}
   \Q_{\Phi}(c_k|\mathbf{X},\mathbf{A}) = \text{softmax}(W^T h_{\Phi, k}), h_{\Phi, i} = \text{GNN}_{\Phi} (x_k, \N_k)
\end{equation}
Now we first describe the process of graph clustering on target graph (uncolored left part in Figure~\ref{fig:framework}). Then, we will introduce how to align those clusters $c \sim \Q_{\Phi}$ with classes on source graph (colored right part in Figure~\ref{fig:framework}).

Given the cluster membership matrix $\mathcal{C} \in \{0,1\}^{|V| \times C} $, the modularity
measure~\cite{newman2006modularity} $\S$ quantifies the divergence between the number of
intra-cluster edges and the expected number of a random graph. By maximizing the modularity, the nodes are densely connected within each cluster:
\begin{equation}
    \S = \frac{1}{2 |E|} \sum_{ij} \left[ A_{ij} - \frac{d_i d_j}{2 |E|} \right] \delta (c_i, c_j)
\end{equation}
where $c_i$ and $c_j$ is the cluster membership of node $i$ and $j$ and $d_i \in \mathbb{R}^{|V|}$ is the degree of the node. Instead of optimizing the binary cluster membership matrix $\mathcal{C}$ (following \cite{tsitsulin2020graph}), our $\text{GNN}_{\Phi}$ optimizes a real-valued cluster assignment matrix $\Q_{\Phi}(y_i|\mathbf{X},\mathbf{A})$ as follows (matrix-form),
\begin{equation}
\label{eq:clustering_alg}
    \mathcal{L}_{\Phi,T} = \frac{1}{2 |E|} \text{Tr} \left( \Q_{\Phi}^\intercal d^\intercal d \Q_{\Phi} - \Q_{\Phi}^\intercal A^t \Q_{\Phi} \right)
\end{equation}
The optimized cluster distribution is the soft cluster membership output by the GNN in Equation~\ref{eq:cluster}.

Given cluster distribution $\Q_{\Phi}^t(c_j|x_j)$ on target graph, we are interested in its correlations with true $\mathbb{P}^t(y_i^t|x_i^t)$. Here we assume a good clustering algorithm works consistently good on source and target \wrt \textit{graph homophily}, that is, $\Q_{\Phi}^s(c_j|x_j) \approx	\mathbb{P}^s(y_i^s|x_i^s) \rightarrow \Q_{\Phi}^t(c_j|x_j) \approx	\mathbb{P}^t(y_i^t|x_i^t)$

To this end, first we need to build a mapping from the graph clusters $C$ to the class labels $Y$ using source graph. 
The number of clusters is set to be no smaller number of classes $N$ ($K \geq N$). Thus, we could build a bipartite mapping between
the clusters $\{c_{k=1}^K\}$ and the classes $\{y_{n=1}^N\}$. We use the KL divergence to measure the
distance between cluster $c_k$ and class $y_n$ over source data.
Specifically, we search for the optimal mapping $\mathcal{M} \in \{0,1\}^{K \times N}$ by solving a linear sum
assignment cost~\cite{kuhn1955hungarian} between them, defined as follows,
\begin{equation}
\label{eq:assignment}
\begin{array}{l}
\min \Sum_k \Sum_n T_{k,n} \mathcal{M}_{k,n}, \\ 
T_{k,n} =  \text{KL}\left(\P_{\theta}(y_n|X,A) || \Q_{\Phi}(c_k|X,A) \right), \\
\text{subject to} \  \forall \Sum_k \mathcal{M}_{k,n} \leq 1, \Sum_n \mathcal{M}_{k,n} \leq 1.
\end{array}
\end{equation}
Thus we can map some clusters to classes, \ie,
$C_L \rightarrow Y, C_L\subseteq	C, |C_L|=N$. After this step, the clusters on source graph are ``colored'' in Figure~\ref{fig:framework}. In order to achieve similar class distribution on source data ($\Q_{\Phi}^s(c_j|x_j) \approx	\mathbb{P}^s(y_i^s|x_i^s)$), we add an adversarial regularization term on clustering,
\begin{equation}
\label{eq:model-q-kl}
    \mathcal{L}_{\Phi, \theta, S} =  \Sum_{i=1}^{B} \text{KL}\left(\mathbb{P}^s(\y|x_i^s,A^s) || \Q_{\Phi}(\y|x_i^s,A^s) \right)
\end{equation}
If labels are all known in source data, we random sample $B$ nodes and $\mathbb{P}^s(y_i^s|x_i^s,A^s)$ is a one-hot vector. Otherwise (semi-supervised), we use current classifier's inference probability on source data $\P_{\theta}(\y|x_i^s,A^s)$.




\subsection{Model Optimization}
\label{sec:co-train}
Following Equation~\ref{eq:open-ssl}, the overall loss for classification is,
\begin{equation}
    \mathcal{L}_{\theta} = \mathcal{L}_{\theta,S} + \mathcal{L}_{\theta,\Phi,T} + \alpha \mathcal{L}_{\theta,\Phi,S}
\end{equation}
where the last term (adversarial samples $(x_i^s, y_i^s)$ on unknown training) is added to ensure training convergence. We have $\alpha=1$ for semi-supervised $\alpha=0$ for full-supervised source data. It is similar with traditional domain adaption~\cite{Ben-David2010} that unlabeled target data involved as $\mathcal{L}_{\theta,\Phi,T}$.

Similarly, the overall loss for adversarial clustering also includes both source and target data,
\begin{equation}
    \mathcal{L}_{\Phi} = \mathcal{L}_{\Phi, T} + \mathcal{L}_{\Phi, \theta, S}
\end{equation}

\xhdr{Joint Optimization.} As we can see in the loss function, both modules require inference result (freezing parameters) from the other. Therefore, we initialize both models with $\mathcal{L}_{\theta, S}$ and $\mathcal{L}_{\Phi, T}$, that is pre-training classification with labeled source graph and clustering with unlabeled target graph. We then iteratively train the classification module and clustering module \textbf{w.r.t.} $\mathcal{L}_{\theta} $ and $\mathcal{L}_{\Phi}$. In $\mathcal{L}_{\theta,\Phi,T}$, we sample same amount of nodes as labeled source data $\D^s$ on target. In adversarial regularization loss $\mathcal{L}_{\Phi, \theta, S}$, we train $\text{GNN}_\Phi$ for $T$ steps and sample $B$ nodes each step. Lastly, Algorithm~\ref{alg} summarizes the joint optimization algorithm.

\xhdr{Convergence.} With the Variational-EM model~\cite{neal1998view}, we now discuss the convergence of graph adversarial clustering. For simplicity, we denote $\mathbf{G} = (\mathbf{X}^s, \text{A}^s)$ as inputs on source data. 
Essentially, we optimize the evidence lower bound (ELBO) of the log-likelihood as follows,
\begin{equation}
\label{eq:VEM}
\begin{array}{l}
\log \P_{\theta}(\y|\mathbf{X}^s, \text{A}^s) \geq \\ \mathbb{E}_{\Q_{\Phi}(\y|\mathbf{X}^s, \text{A}^s) } \left[ \log \P_{\theta}(\y|\mathbf{X}^s, \text{A}^s) -\log \Q_{\Phi}(\y|\mathbf{X}^s, \text{A}^s) \right]
\end{array}
\end{equation}
In t-th E-step, regarding the Equation~\ref{eq:VEM}, the optimal $\Q_\Phi^{(t+1)}$ is,
\begin{equation}
\begin{array}{l}
    \log \Q_{\Phi}^{(t+1)}(\y|\mathbf{G}) = \mathbb{E}_{\Q_{\Phi}^{(t)}(\y|\mathbf{G}) } \left[ \log \P_{\theta}^{(t)}(\y|\mathbf{G})\right] + \text{const}
\end{array}
\end{equation}
The solution of the above E-step is $\Q_{\Phi}^{(t+1)}(\y|\mathbf{G}) = \argmin \text{KL}(\P_{\theta}^{(t)}(\y|\mathbf{G})||Q_{\Phi}^{(t)}(\y|\mathbf{G}))$. In practice, we achieve this update by sample $B$ nodes on source graph, which is exactly the adversarial regularization term (Equation~\ref{eq:model-q-kl}).

In t-th M-step, we update the output distribution on source from classifier $\P_{\theta}^{(t+1)}$ with $\Q_{\Phi}^{(t+1)}(\y|\mathbf{G})$.
To estimate the expectation term in Equation~\ref{eq:VEM}, we sample
$(x_j^s,y_j^s) \sim \Q_{\Phi}$ to compute the log-likelihood and update parameter $\theta$ of the classifier. Given the alignment between cluster and classes in Section~\ref{sec:model-q}.
Interestingly, we describe the same sampling process on target in $\mathcal{L}_{\theta,\Phi,T}$ and now we add $\mathcal{L}_{\theta,\Phi,S}$ in final loss of classifier to accomplish the M-step. Therefore, at each episode, the joint optimization is equivalent to perform variational EM alternatively. The convergence is proven in the original paper~\cite{neal1998view}.

\xhdr{Complexity.}
Compared with normal node classification, our extra computation comes from graph adversarial clustering. Hereby, we analyze the additional computation cost in $\Q_{\Phi}$.  Let $\mathcal{O}(\Phi)$ be the time GNN $\Phi$ takes to compute a single node embedding and $\mathcal{T}$ for compute cluster membership for each node. Assuming there are $|V^t|$ nodes in target graph, the pre-training of clustering takes $\mathcal{O}(|V^t|\cdot (\Phi+\mathcal{T}))$. In addition, the adversarial regularization costs $\mathcal{O}(2\cdot \mathcal{E} \cdot B \cdot \Phi )$, $\mathcal{E}$ is the number of episode takes the algorithm to converge. Overall, the extra complexity is in linear of target graph size, which is reasonable for domain adaption.

\begin{algorithm}[ht]
 /* Episode 0 */ \\
 Pre-train the classifier $\text{GNN}_\theta$ via $\mathcal{L}_{\theta, S}$ and the cluster $\text{GNN}_\Phi$ via $\mathcal{L}_{\Phi, T}$\;
 
 /* Iterative optimization, episode 1,2...*/ \\
 \While{Micro-F1 on validation increases}{
  /* update $\text{GNN}_\Phi$ for T steps each episode */\\
\For{t = 1 to $T$}{
\text{Sample} $B$ nodes $(x_i^s, y_i^s)$ from source\\
 $\Phi \xleftarrow{-} \nabla_{\Phi} \mathcal{L}_{\Phi,S} + \nabla_{\Phi} \mathcal{L}_{\Phi,\theta,T}$
 }
 /* update shift-robust classifier $\text{GNN}_\theta$ */\\
\text{Sample} $|\D^s|$ nodes $(x_j^t, y_j^t)$ from target\\
 \While{$\mathcal{L}_{\theta}$ not converge} {
 $\theta \xleftarrow{-} \nabla_{\theta} \mathcal{L}_{\theta,S} + \nabla_{\theta} \mathcal{L}_{\theta,\Phi,T} + \nabla_{\theta} \alpha \mathcal{L}_{\theta,\Phi,S}$
}

 }
 \caption{Pseudo code for \Ours optimization}
 \label{alg}
\end{algorithm}
\section{Experiments}
\label{sec:experiments}

\begin{table*}[h]
\caption{Open-set classification on three different citation networks. Numbers reported are all percentage (\%). Different graph neural networks in second block all uses progressive pseudo-labeling(previous SoTA on open-set shift).}
\label{tab:classification-result}
\scalebox{0.95}{
\begin{tabular}{ll|c|c|c|c|c|c|c|c|c}
\toprule
\multicolumn{2}{c|}{\multirow{2}{*}{Method}}                             & \multicolumn{3}{c|}{\textbf{Cora}}            & \multicolumn{3}{c|}{\textbf{Citeseer}} & \multicolumn{3}{c}{\textbf{PubMed}}                                                                                                                      \\
\multicolumn{2}{c|}{}                                                    & Micro-F1$\uparrow$ & Macro-F1$\uparrow$ & $\Delta$F1$\downarrow$ & Micro-F1$\uparrow$ & Macro-F1$\uparrow$ & $\Delta$F1$\downarrow$ & Micro-F1$\uparrow$ & Macro-F1$\uparrow$ & $\Delta$F1$\downarrow$ \\ \midrule
\multicolumn{2}{l}{GCN (IID)}                  &  80.8 $\pm$ 0.4       &  79.8 $\pm$ 0.3    &  0               &  71.1 $\pm$ 0.5                &  68.3 $\pm$ 0.5      &  0     &  79.3 $\pm$ 0.3  & 78.8 $\pm$ 0.3    &       0                 \\
\midrule
\multicolumn{2}{l}{PGL+DGI}                    &   70.2 $\pm$ 4.2       &    68.9 $\pm$ 4.0    & 10.6              &  62.9$\pm$ 5.1    & 56.4 $\pm$ 10.7        &  7.4   & 58.8 $\pm$ 6.1 & 47.0 $\pm$ 2.0 & 20.5 \\ 
\multicolumn{2}{l}{PGL+GCN}                  &  72.1 $\pm$ 4.4       &   70.9 $\pm$ 4.8    &  8.7              &  67.0 $\pm$ 5.2    &  60.0 $\pm$ 9.4       &  14.1     & 63.6 $\pm$ 3.8 & 57.8 $\pm$ 7.0   & 15.7                            \\
\multicolumn{2}{l}{PGL+GAT}          & 69.2 $\pm$ 5.0        &   67.4 $\pm$ 5.2   & 11.5 &  65.4 $\pm$ 4.9               &  55.4 $\pm$ 10.1         &  15.7    &  63.1 $\pm$ 4.0 &  57.3 $\pm$ 6.2  & 16.2  \\
\multicolumn{2}{l}{PGL+SAGE}                 &  73.3 $\pm$ 4.3       &   71.8 $\pm$ 5.0    &  7.5        &  67.2 $\pm$ 5.4    & 60.2 $\pm$ 8.8        & 3.9  & 66.1 $\pm$ 1.7 & 60.4 $\pm$ 8.1 & 13.2 \\ 
\multicolumn{2}{l}{PGL+SGC}                   &  68.9 $\pm$ 5.0       &   66.9 $\pm$ 6.1    &  11.9             &  62.3 $\pm$ 7.0    &  56.4 $\pm$ 10.1       &  8.8   &  63.3 $\pm$ 4.3   &   57.2 $\pm$ 6.4   &        16.0                    \\ \midrule
\multicolumn{2}{l}{\Ours \textbf{w.o} $\Phi$}               &  71.7 $\pm$ 6.4       &   70.2 $\pm$ 3.6     & 9.1                &  65.5 $\pm$ 4.7   & 56.2 $\pm$ 4.5       &   5.6    &  65.8 $\pm$ 1.6                &  60.5 $\pm$ 7.4 &  13.5 \\
\multicolumn{2}{l}{\Ours Ep.1}               &  \underline{76.0 $\pm$ 4.7}       &   \underline{75.2 $\pm$ 2.9}     & \underline{4.8}               &  \underline{69.2} $\pm$ \underline{3.0}    &  \underline{60.4 $\pm$ 6.0}       &   \underline{1.9}    &  \underline{67.3 $\pm$ 5.1}                &  \underline{68.0  $\pm$ 3.9} &  \underline{12.0}  \\
\multicolumn{2}{l}{\Ours}                 &  \textbf{77.4 $\pm$ 4.0}       &   \textbf{75.9 $\pm$ 3.6}     & \textbf{3.4}   &  \textbf{70.7 $\pm$ 4.0}    &  \textbf{63.4 $\pm$ 7.4}       &   \textbf{0.4}    &  \textbf{69.1 $\pm$ 4.4}   &  \textbf{69.4 $\pm$ 2.5}  & \textbf{10.2} \\ \midrule 
\bottomrule

\end{tabular}
}
\end{table*}

In this section, we will empirically evaluate
\Ours. We focus on answering the following questions in our experiments:

\begin{itemize}
    \item Can \Ours perform well on on open-set distribution shift ?
    \item Does \Ours outperform existing domain adaption methods on close-set shift on dynamic graph ?
\end{itemize}

\subsection{Experimental Setting}

\begin{table}[]
    \caption{Overall Dataset Statistics}
    \label{tab:dataset-stats}
    \centering
    \begin{tabular}{c|c c c c }
    \toprule
         Dataset & \# Nodes & \# Edges & \# Classes & \# Train\\
         \midrule
         Cora & 2,708 & 5,429  & 7  & 140 \\
         Citeseer & 3,327  &   4,732 & 6 & 120 \\
         PubMed & 19,717  &  44,325 & 3 & 100 \\
         ogb-arxiv & 169,343  &   2,501,829 & 40 & 12974 \\
         \bottomrule
    \end{tabular}
\end{table}
\xhdr{Datasets.}
In our experiments, we perform node classification tasks on four benchmark networks (see Table~\ref{tab:dataset-stats}),
These four networks are: Cora, Citeseer, PubMed~\cite{sen2008collective} and ogb-arxiv~\cite{hu2020open}. We conduct open-set shift experiments on first three datasets and close-set shift on ogb-arxiv, because ogb-arxiv already exhibits close-set shift across different time periods.


\xhdr{Compared algorithms.} 
We compare \Ours with strong baselines on open-set shift and close-set shift. 
Since many domain adaption algorithms cannot deal with open-set shift, we apply the previous state-of-the-art open-set progressive graph learning (PGL)~\cite{luo2020progressive} on different GNNs such as GCN~\cite{kipf2016semi}, GraphSAGE~\cite{hamilton2017inductive}, SGC~\cite{wu2019simplifying}, GAT~\cite{velivckovic2017graph}, DGI~\cite{velivckovic2018deep}. On close-set shift, we provide the result from multiple close-set domain adaption methods on two different GNNs: (1) GCN (2) DGI~\cite{velivckovic2018deep}. These close-set domain adaption methods are as follows,
\begin{enumerate}
    \item DANN~\cite{ganin2016domain} minimizes the feature representation discrepancy by an adversarial domain classifier. We add the adversarial head at the last hidden layer with activation in GCN and MLP.
    \item CMD~\cite{zellinger2017central} matches the means and higher order moments between source and target in the latent space.
    \item SRGNN~\cite{zhu2021shift} is a recent shift-robust algorithm for localized training data, which combines CMD and importance sampling to improve the generalization on target.
\end{enumerate}
To provide a comprehensive study of \Ours, we design two ablations of our algorithms:
\begin{itemize}
\item \Ours \textbf{w.o} $\Phi$: a bootstrapping ablation, where we pseudo label the same amount of data on target using model $\theta$ predictions instead of adversarial clustering $\Phi$ in Equation~\ref{eq:open-ssl}.
\item \Ours Ep.1: we stop the algorithm after episode 1 to verify the effectiveness of iterative optimization.
\end{itemize}

\xhdr{Evaluation Metrics.} 
In our experiments,
we have $\{(x^t,y^t)\}$ in testing (target) data $\mathcal{D}^t$ and calculate Micro-F1 and Macro-F1 on predictions $\hat{y}$ from target data,
\begin{align}
    \text{Micro-F1} = \frac{| x\in D^t \wedge	\hat{y}=y^t | }{|(x,y) \in D^t|} \\
    \text{Macro-F1} = \frac{1}{N_\text{test}}\Sum_{i=1}^ {N_\text{test}} \frac{| x\in D^t \wedge	\hat{y}=y^t | }{| x \in D^t\wedge y^t=i|}
\end{align}
Notice that in open-set shift, target data $\D^t$ has one more class than training - the unknown class ($N_\text{test} = N_\text{train} + 1)$).

\xhdr{Parameter Settings and Reproducibility.}
In \Ours, we use the  GCN~\cite{kipf2016semi} and add self-loop to every node in the graph.
We use the DGL~\cite{wang2019dgl} implementation of different GNN architectures. The Adam SGD~\cite{kingma2014adam} optimizer is used for training with
learning rate as 0.01 and weight decay as 0.0005. The hidden size of all GNN models including \Ours
is 256. We set cluster number as 16 in cluster $\text{GNN}_\Phi$ for open-set experiment and same as number of classes in close-set experiment. Note that none of Cora, Citeseer and Pubmed has 16 classes, which do not favor our method as a fair comparison.

\begin{table}[h]
\caption{Close-set Shift classification on ogb-arxiv.}
\label{tab:close-set-result}
\centering
\scalebox{0.93}{
\begin{tabular}{ll|c|c|c}
\toprule
\multicolumn{2}{c|}{\multirow{2}{*}{Method}}                             & \multicolumn{3}{c}{\textbf{ogb-arxiv}}                                                                                                                               \\
\multicolumn{2}{c|}{}                                                    & 2014-2016 & 2016-2018 & 2018-2020 \\ \midrule
\multicolumn{2}{l}{GCN} &  56.2 $\pm$ 0.5       &   55.7 $\pm$ 0.8     &  53.8 $\pm$ 1.2                 \\

\multicolumn{2}{l}{GCN-DANN} &  54.3 $\pm$ 1.0       &   50.4 $\pm$ 3.2    &  46.2 $\pm$ 5.0                   \\
\multicolumn{2}{l}{GCN-CMD} & 50.7 $\pm$ 0.6       &   48.7 $\pm$ 1.5    &  50.0 $\pm$ 2.3             \\
\multicolumn{2}{l}{GCN-SRGNN} & 54.4 $\pm$ 0.6       &   53.3 $\pm$ 1.1   &  55.0 $\pm$ 1.1             \\
\midrule
\multicolumn{2}{l}{DGI} &  52.6 $\pm$ 0.4       &   48.3 $\pm$ 1.9    &  50.9 $\pm$ 1.4                   \\
\multicolumn{2}{l}{DGI-DANN} &  48.9 $\pm$ 1.5       &   44.4 $\pm$ 3.1    &  28.2 $\pm$ 0.7                   \\

\multicolumn{2}{l}{DGI-CMD} &  44.5 $\pm$ 0.6       &   36.5 $\pm$ 1.0    &  31.0 $\pm$ 1.9                   \\
\multicolumn{2}{l}{DGI-SRGNN} & 50.5 $\pm$ 1.8       &   49.7 $\pm$ 2.7    &  47.7 $\pm$ 2.2             \\
\midrule 
\multicolumn{2}{l}{\Ours \textbf{w.o} $\Phi$}  & \underline{57.3 $\pm$ 0.2 }      &    \underline{58.0 $\pm$ 0.8}     &   \underline{55.6 $\pm$ 1.7}              \\
\multicolumn{2}{l}{\Ours Ep.1}  &  56.9 $\pm$ 0.1       &   56.0 $\pm$ 0.4     & 54.5 $\pm$ 0.1              \\
\multicolumn{2}{l}{\Ours} &  \textbf{58.1 $\pm$ 0.3}       &   \textbf{58.7 $\pm$ 0.8}     &  \textbf{59.1 $\pm$ 1.3} \\
\bottomrule
\end{tabular}
}
\vspace{-1em}
\end{table}

\subsection{Open-Set Shift with Unseen Classes}
\label{sec:exp-results}

We create synthetic open-set shift by removing 3 classes on Cora/Citeseer and one class on PubMed from training data. In testing, nodes from the masked classes are all re-labeled as the unknown class.
For each known class, we random sample 20 nodes for training and report the mean and standard deviation of micro-F1 and macro-F1 for 10 runs. Besides, we report the performance of a GCN with full class visibility - GCN(IID) in Table~\ref{tab:classification-result}, and relative performance drop (of other methods) in Micro ($\Delta\text{F1}$). 
In validation set, we have nodes from the unknown class and only use it to select the best hyper parameters for both PGL~\cite{luo2020progressive} and \Ours. For example, in PGL, we use validation to pick the best threshold $\alpha$ among each episodes $\{\alpha_k\}$ and label nodes with lower probabilities into unknown class. \Ours has an explicit class for unknown, so we use the Micro-F1 on validation to select batch size B and step size T for $\text{GNN}_\Phi$.

On average,
\Ours  outperforms all the other representative GNNs + PGL for at lease 4$\%$ and 2$\%$ in micro-F1 and macro-F1, respectively.
Among the baselines, the most competitive baseline is GraphSAGE+PGL. It is probably because the self information is most sensitive to distributional shifts and GraphSAGE specifically promotes self information. 
PubMed is the largest graph among three datasets and open-set shift is larger.
The experiment also shows that the end-to-end supervised GNNs (GCN) are
more robust than unsupervised GNNs (DGI) when there is open-set shift. 


From the comparison between our own ablations, we find that the graph adversarial clustering (see \Ours \textbf{w.o.} $\Phi$) contributes the most to the performance gains. This ablation randomly samples pseudo unknown nodes from low confident predictions.
In other words, drawing samples
from unseen clusters is the key towards better open-set generalization. Moreover, the
iterative brings more improvement (1$\sim$3\% F1) on top of \Ours Ep.1 ablation since variational EM takes several round to converge in practice.

\subsection{Close-Set Shift on Dynamic Graph}
\label{sec:close-set}
On ogb-arxiv, we split the train, validation and test nodes by year. There are total 12974 paper before 2011 (train), 28151 (validation), 28374 (2014-2016), 51241 (2016-2018), 48603 (2018-2020). 
In testing, each test graph contains all previous nodes such that train graph is a subset of all test graphs. This dynamic graph setting is close to real world applications, where model is trained on an earlier snapshot (source) and deployed on real-time graphs (target).

First, we are wondering whether unsupervised representation learning is more robust to distribution shifts.
We choose DGI as unsupervised GNN approach and apply DGI encoder optimized on test graph to obtain training and testing node embeddings. From Table~\ref{tab:close-set-result}, we can observe the performance of DGI is worse than GCN, which is consistent with open-set shift - unsupervised GNNs are more vulnerable to distributional shifts. 
Most of exiting close-set domain adaption baselines, focus on learning invariant feature representations between source and target. However, as we stressed before, conditional shifts widely exist in real-world graphs so regularization (CMD) and adversarial head (DANN) make the performance even worse compared with original model.
The performance of \Ours is significantly better than other baselines and base model in different test period because we are able to capture the joint distribution $\Pr_\text{test}(X,Y)$ on target data. The results suggest that our proposed method is more robust in real-world close-set shift. The bootstrapped pseudo labels (\Ours \textbf{w.o.} $\Phi$) from classifier can only marginally improve the performance and the improvement on GCN is smaller under larger shifts (2018-2020 split). \Ours Ep.1 also reports worse results since the algorithm is not converged yet. Our results demonstrate that our framework is effective under different close-set shift introduced by dynamic graph.

\subsection{Analysis and Use Case}
\label{sec:analysis}

\begin{table}[]
\caption{Open-set performance improvement when number of cluster changes from 16  to 7 (best hyper-parameter) on Cora.}
\label{tab:parameter-sensitivity}
\centering
\scalebox{0.9}{
\begin{tabular}{c|cc|cc}
\toprule
{\multirow{2}{*}{}}                             & \multicolumn{2}{c|}{Cora (1 unseen class)} & \multicolumn{2}{c}{Cora (3 unseen classes)} \\
& \multicolumn{1}{c|}{Micro-F1} & Macro-F1 & \multicolumn{1}{c|}{Micro-F1}             & Macro-F1            \\ \midrule
$\Delta$        & \multicolumn{1}{c|}{+$4.73$}  & +$3.79$   &  \multicolumn{1}{c|}{+$1.62$}  & {+$1.15$}     \\\bottomrule
\end{tabular}
}
\vspace{-1em}
\end{table}

\xhdr{Parameter Sensitivity.} Among all the hyper-parameters, the pre-defined number of clusters $|C|$ seems to be the most crucial in open-set scenario.
Now we provide the performance of \Ours with varied numbers of clusters $k$,
\ie 7 (optimal) vs. 16 (used in experiment) in Table~\ref{tab:parameter-sensitivity}. Both Micro-F1 and Macro-F1 are further boosted, since seven is the ground truth number of classes. On the one hand, \Ours is not sensitive to the choice of $|C|$ and better than baselines regardless, especially when there are multiple missing classes (closer to the real-world setting). 
On the other, a accurate estimation of $|C|$ can facilitate the model training. 

\xhdr{Use Case.} Our model do not have specific requirement on classification GNN architectures. Hence, when there are unseen class or substantial close-set shifts expected in testing, our graph adversarial clustering can be plugged in and joint optimized with convergence guarantee. The clustering output can help find potential unseen classes and we provide a case study on DBLP academia network in Appendix.

\section{Conclusion}
In this paper, we proposed a general framework, \Ours, to enable graph neural networks for distributional shift. 
Different from existing work on either \textit{open-set shift} or \textit{close-set shift}, \Ours works for both scenario with a novel graph adversarial clustering component.
Accordingly, \Ours employ a latent variable model (\ie cluster
GNN) to model the latent structure and allows the clustering GNN to improve the generalization of shift-robust
classification GNN.
Future work includes applying the model to applications such as medical diagnosis and molecular dynamics, where robustness towards distribution shifts are in critical need.

\bibliography{ref}
\bibliographystyle{abbrv}

\newpage
\appendix

\section{Appendix}
\subsection{Additional Experiments for Clustering GNN $\Q_\Phi$}
In \Ours,  adversarial regularization term on clustering (Equation~\ref{eq:model-q-kl}) also calibrates the cluster $\text{GNN}_{\Phi}$. The quality of the clustering is a key step towards successfully open-set classification. Therefore, in Table.~\ref{tab:zeroshot-result},
we report the clustering performance on two data sets against Constrained-KMeans with different node embeddings.
Three classes are
masked during the training of \Ours.
The superior performance of \Ours on ACC and NMI validates that
the adversarial regularization on clustering also helps generalization on unseen labels. In particular, without the $\mathcal{L}_{\Phi, \theta, S}$, our cluster GNN (modularity based) reports similar numbers with DGI as both methods are based on GNN.

\begin{table}[h]
\caption{Performance of unseen class generalization on Cora and Citeseer. We utilize constrained-KMeans for baselines.}
\label{tab:zeroshot-result}
\scalebox{1.0}{
\begin{tabular}{l|cc|cc}
\toprule
{\multirow{2}{*}{Method}}                             & \multicolumn{2}{c|}{Cora} & \multicolumn{2}{c}{Citeseer} \\
& \multicolumn{1}{c|}{ACC} & NMI & \multicolumn{1}{c|}{ACC}             & NMI            \\ \midrule
Node features        & \multicolumn{1}{c|}{33.80}      &  13.70  &       \multicolumn{1}{c|}{28.80}            &   11.85            \\
\midrule
 DeepWalk                  & \multicolumn{1}{c|}{47.70}      &  37.26  &       \multicolumn{1}{c|}{22.48}            &   27.63            \\
 node2vec            & \multicolumn{1}{c|}{48.90}      &  40.24  &       \multicolumn{1}{c|}{21.81}            &   27.81            \\
 DGI                 &  \multicolumn{1}{c|}{70.56}      &  51.93  &       \multicolumn{1}{c|}{37.68}            &   42.20            \\
 \midrule
 \Ours \textbf{w.o.} $\mathcal{L}_{\Phi, \theta, S}$                 & \multicolumn{1}{c|}{70.20}      &  53.33  &       \multicolumn{1}{c|}{43.50}            &   42.09            \\
\Ours                 & \multicolumn{1}{c|}{ \textbf{74.25}}      &  \textbf{56.87}  &       \multicolumn{1}{c|}{\textbf{49.16}}            &   \textbf{46.22}            \\\bottomrule
\end{tabular}
}
\end{table}

\subsection{Additional Experiments for Open-set Application}
\xhdr{Dataset.} DBLP is a computer science bibliography website. We construct a heterogeneous network upon three types of nodes:
\textbf{(A)}Author, \textbf{(P)}Paper and \textbf{(T)}Term. There are five types of papers labeled
by the corresponding venues\footnote{\textbf{DM}: SIGKDD, ICDM\ \textbf{DB}: SIGMOD, VLDB\
  \textbf{NLP}:ACL, EMNLP, NAACL\ \textbf{AI}: AAAI, IJCAI, ICML\ \textbf{CV}: ECCV, ICCV, CVPR} -
Data Mining \textbf{(DM)}, Database \textbf{(DB)}, Artificial Intelligence \textbf{(AI)}, Natural
Language Processing \textbf{(NLP)} and Computer Vision \textbf{(CV)}. We sampled 50,000 papers
between 2010 and 2015, then use the venue information to label 5567 of them into 5 classes. After
filtering out rare authors and terms, there are 20601 papers, 5158 authors and 2000 terms in
total. In DBLP, we have more than 70 \% of 20,000 papers as instances from  unseen classes. In other
words, the majority of nodes in the DBLP graph belong to unseen classes.

\xhdr{Unseen Class Discovery.} In main paper, we describe the use case of \Ours to discover the new classes and facilitate the design of label space. 
In Table~\ref{tab:case-supp}, we show the paper in DBLP dataset that are assigned to the  (unmatched) clusters, discovered by our Cluster GNN $\Q_\Phi$ and
Constrained-KMeans + DGI embeddings.
Interestingly, we find two shared clusters about ``Computational Mathematics'' and ``Network \&
Distributed System'' from two algorithms. The results show ad-hoc unsupervised clustering on node
embeddings produces erroneous cluster membership even for the top-ranked items. For example, it contains
almost half papers from the computer architecture in ``Network \&
Distributed System'', whereas unseen classes suggested by our
adversarial regularized cluster are much more consistent.

\xhdr{Effectiveness of iterative optimization.} In Figure~\ref{fig:calibration}, we demonstrate the
learning procedure on DBLP dataset. Each column shows the predicted labels from classification GNN
(upper) and the  cluster assignments from the cluster GNN (lower). We visualize the node embeddings
from cluster GNN into two dimensions using t-SNE. In this example, we
set number of clusters to six. In Figure~\ref{fig:classification-a} and \ref{fig:cluster-a}, the
pre-trained classification GNN predicts all papers into 5 seen classes. Our cluster GNN shows 6
initial clusters. During the training, the clusters in Figure~\ref{fig:cluster-b} gradually
align with the  classifier's predictions in same color. The classifier presented in Figure~\ref{fig:classification-b} starts to recognize nodes of unseen class with samples from cluster corresponds to unseen class (yellow) in Figure~\ref{fig:cluster-b}.
When the training converges,  the clusters in Figure~\ref{fig:cluster-c} are separated clearly.

\subsection{Visualization of Close-set Shifts}
\begin{figure}[h]
    \centering
    \includegraphics[width=0.43\textwidth]{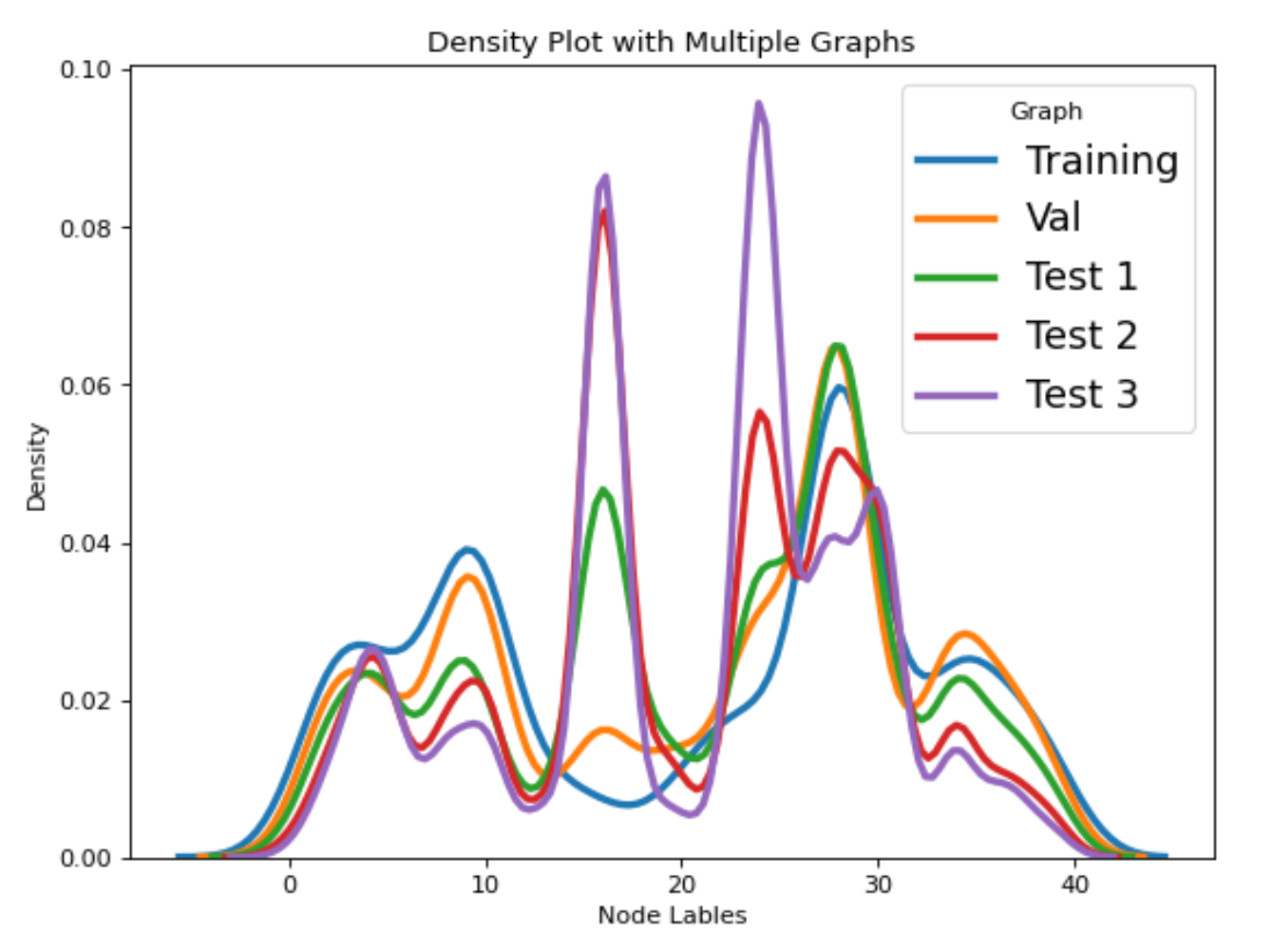}
    \caption{Label shift across different testing-time on ogb-arxiv.}
    \label{fig:case-close-set}
\end{figure}

\begin{figure*}
\centering
        \begin{subfigure}[b]{0.3\textwidth}
            \centering
            \includegraphics[width=\textwidth]{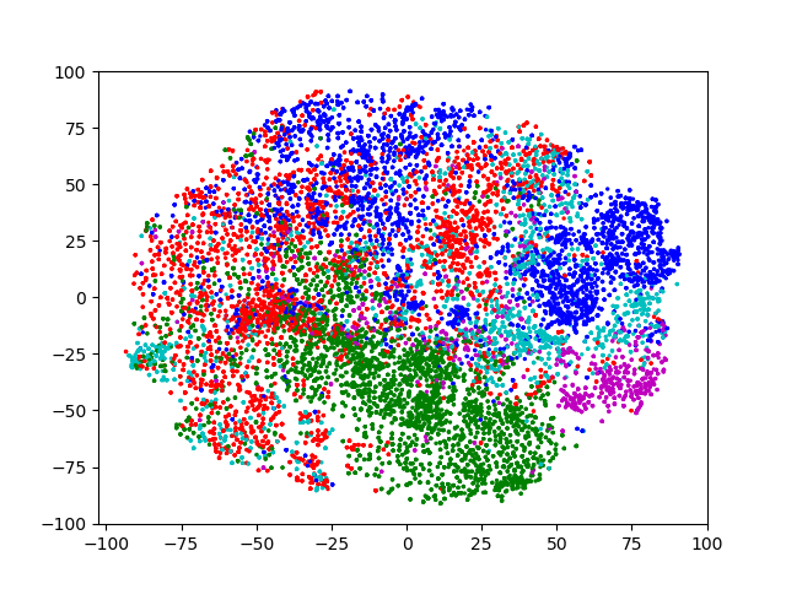}
            \caption{pre-train done, no unseen class}
            \label{fig:classification-a}
        \end{subfigure}
     \begin{subfigure}[b]{0.3\textwidth}
             \centering
             \includegraphics[width=\textwidth]{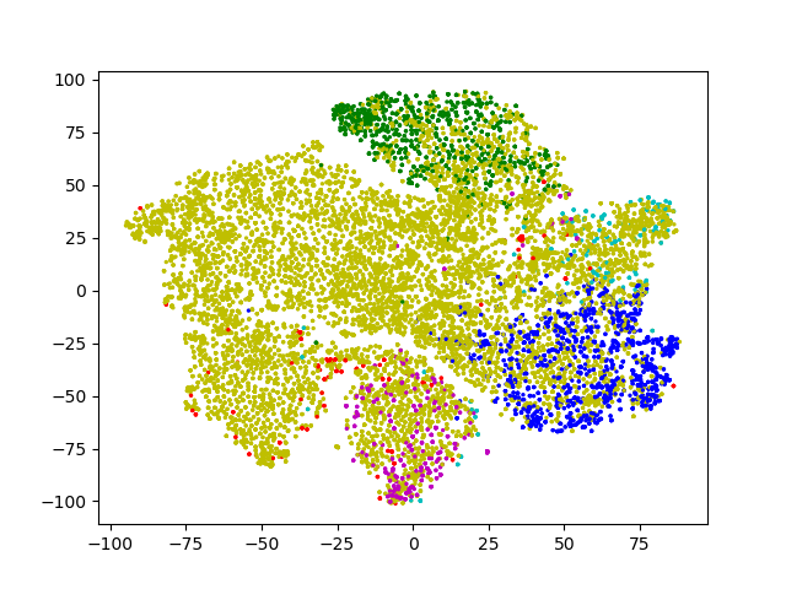}
             \caption{discover unseen during training}
             \label{fig:classification-b}
     \end{subfigure}
          \begin{subfigure}[b]{0.3\textwidth}
             \centering
             \includegraphics[width=\textwidth]{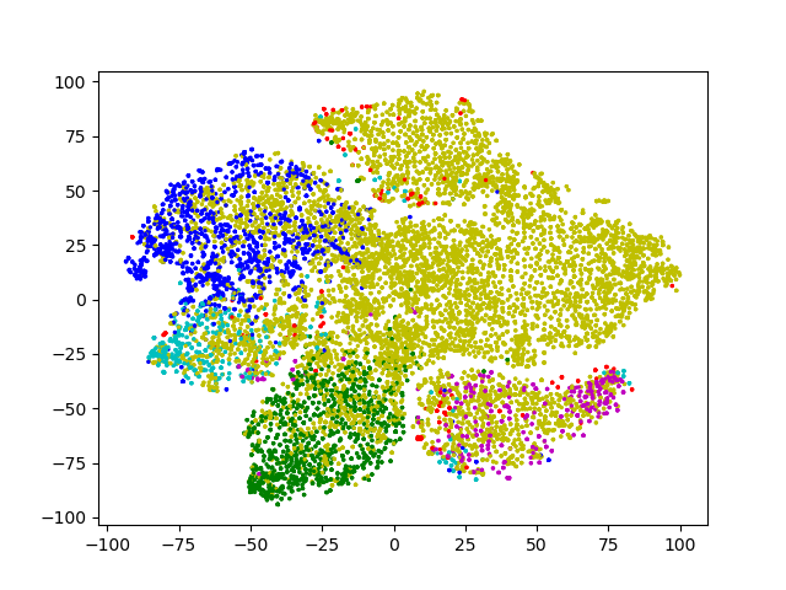}
             \caption{classifier component converges}
             \label{fig:classification-c}
     \end{subfigure}

             \begin{subfigure}[b]{0.3\textwidth}
            \centering
            \includegraphics[width=\textwidth]{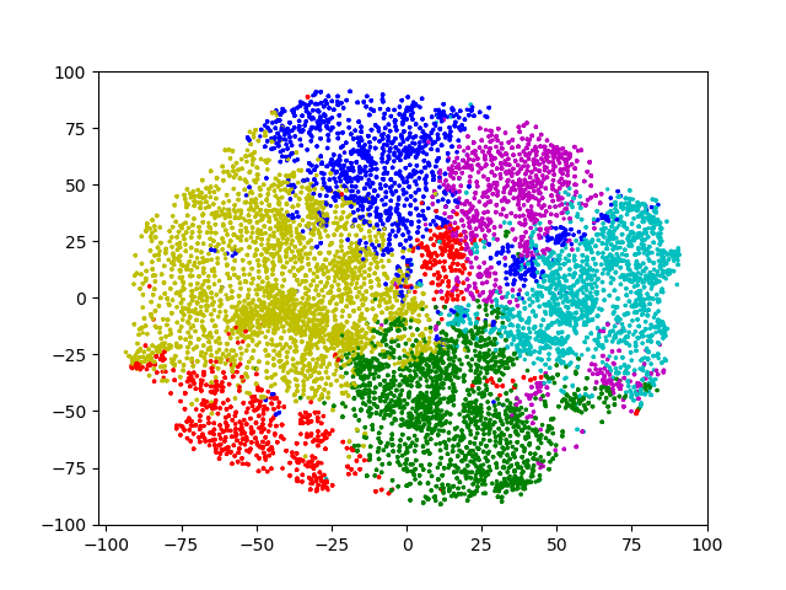}
            \caption{pre-train, no adversarial regularization}
            \label{fig:cluster-a}
        \end{subfigure}
     \begin{subfigure}[b]{0.3\textwidth}
             \centering
             \includegraphics[width=\textwidth]{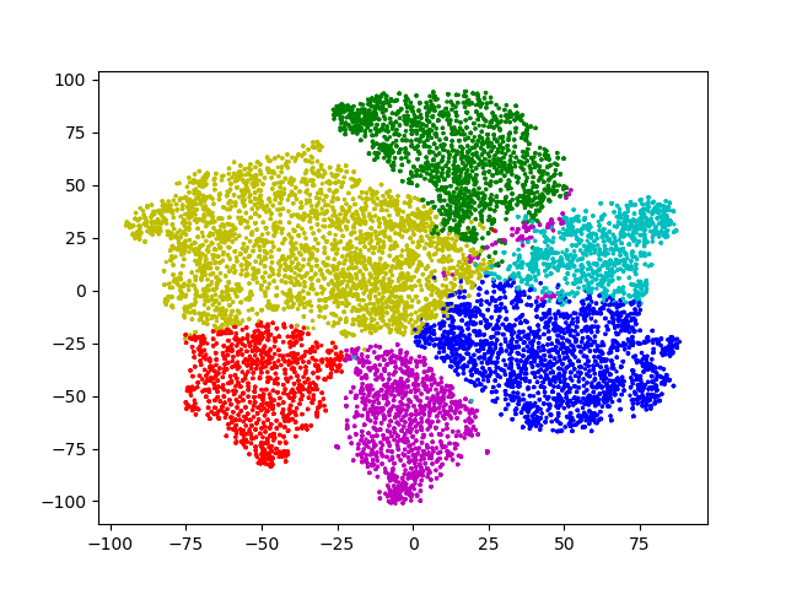}
             \caption{distinct cluster in adversarial training}
             \label{fig:cluster-b}
     \end{subfigure}
          \begin{subfigure}[b]{0.3\textwidth}
             \centering
             \includegraphics[width=\textwidth]{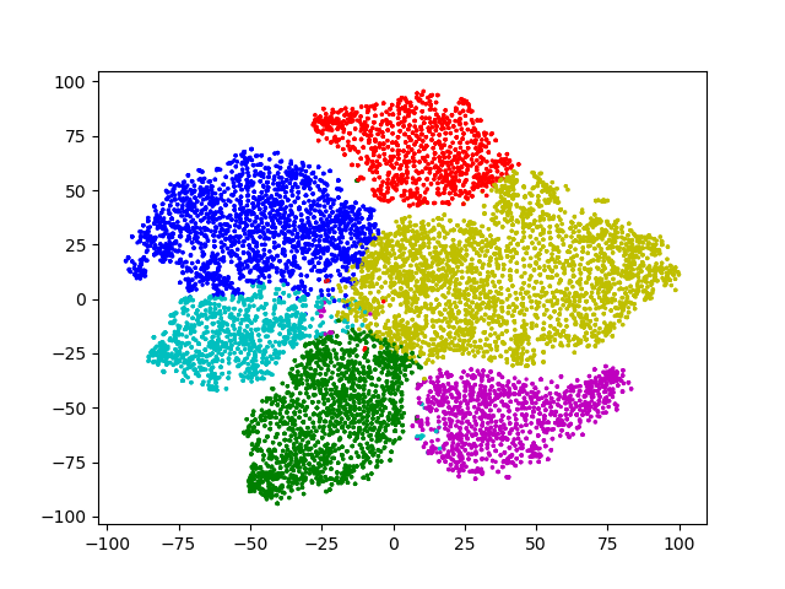}
             \caption{cluster component converges}
             \label{fig:cluster-c}
     \end{subfigure}
    \caption{Visualization of training process on DBLP dataset (Best viewed in color). The same color indicates the aligned class/cluster (Red: AI, Blue: CV, Cyan: DB, Green: DM, Purple: NLP). Color yellow represents the unseen class detected by the classifier.}
    \label{fig:calibration}
\end{figure*}

In Section~\ref{sec:close-set} of the main paper, different algorithms suffer from close-set shift between training and testing-time graphs. Although it would be hard to directly demonstrate the conditional shift on $\Pr(Y|X)$, we provide the label distribution on 40 classes in Figure~\ref{fig:case-close-set}. As we can observe, the distribution changes dramatically betweeen train/val and three test splits. Thus, a shift-robust classification algorithm is in a great need for real-world node classifications.

\begin{table*}
\centering
 \caption{Top paper titles from unmatched clusters on DBLP dataset. We choose the two most similar clusters for comparison. \textbf{(x)} indicates the paper is not consistent with the area of other papers in the same cluster.} 
 \label{tab:case-supp}
\scalebox{0.7}{
\begin{tabular}{c|l||l}
\toprule
    Unseen Cluster & \multicolumn{1}{c}{\textbf{Cluster GNN} } &  \multicolumn{1}{c}{\textbf{Constrained-KMeans (DGI)}}  \\ \midrule
      \multirow{20}{*}{\shortstack[c]{\textbf{Computational} \\ \textbf{Mathematics} }} & \textbullet\ a meshless method for solving nonlinear two-dimensional   & \textbullet\ upper and lower bounds for dynamic cluster assignment 
\\
     & ... domains using radial basis functions with error analysis & for multi-target tracking in heterogeneous wsns\textbf{(x)} \\
      & \textbullet\ application of the ritz-galerkin method ... wave equation & \textbullet\ multi-rate multicast ... multi-radio wireless mesh networks\textbf{(x)}\\
      & \textbullet\ a generalized moving least square reproducing kernel method  & \textbullet\ matrix equations over (r,s)-symmetric and ... symmetric matrices\\
      & \textbullet\  multi-rate multicast ... multi-radio wireless mesh networks \textbf{(x)} & \textbullet\ a tau approach for solution of the ... fractional diffusion equation\\
      & \textbullet\ numerical solution of the higher-order ... variable coefficients & \textbullet\ a meshless method for solving nonlinear two-dimensional \\ 
      & equation with variable coefficients & ... domains using radial basis functions with error analysis \\ 
      & \textbullet\ on a system of difference ... odd order solvable in closed form  & \textbullet\ two class of synchronous matrix ... linear complementarity problems \\
      & \textbullet\ two class of synchronous matrix ... complementarity problems & \textbullet\ a generalized moving least square reproducing kernel method\\
      & \textbullet\ upper and lower bounds for dynamic ... heterogeneous wsns& \textbullet\ application of the ritz-galerkin method...in the wave equation\\
      & \textbullet\ a finite volume spectral element ... (mhd) equations & \textbullet\ a finite volume spectral element ... (mhd) equations\\
      & \textbullet\ a tau approach for solution ... space fractional diffusion equation & \textbullet\ numerical solution of the higher-order linear ... variable coefficients\\
      & \textbullet\ numerical solution ... using chebyshev cardinal functions & \textbullet\ a technique for the numerical ... birkhoff-type interpolation method\\
      & \textbullet\ computation of two time-dependent ...additional specifications & \textbullet\ computation of two time-dependent ... additional specifications\\
      & \textbullet\ a technique for the ... birkhoff-type interpolation method & \textbullet\ numerical solution ... using chebyshev cardinal functions\\
      & \textbullet\ matrix equations over (r,s)-symmetric ...symmetric matrices & \textbullet\ weight evaluation ... constrained data-pairscan't-linkq\\
      & \textbullet\ on a solvable system of difference equations of kth order & \textbullet\ single-machine scheduling ... delivery times and release times \textbf{(x)}\\
      & \textbullet\ on an integral-type operator from w-bloch ... u-zygmund spaces & \textbullet\ new results ... past-sequence-dependent delivery times \textbf{(x)}\\
      & \textbullet\ on an integral-type ... to mixed-norm spaces on the unit ball & \textbullet\ adaptive finite element ... pointwise and integral control constraints\\
      & \textbullet\ composition operators ... on the unit disk and the half-plane & \textbullet\ sparsely connected neural network-based time series  forecasting \textbf{(x)}\\
      & \textbullet\ norms of some operators on bounded symmetric domains & \textbullet\ inexact solution of nlp subproblems in minlp\\
      & \textbullet\ on a solvable system of difference equations of fourth order & \textbullet\ some goldstein's type methods ... variant variational inequalities\\

      \midrule
      
      \multirow{20}{*}{\shortstack[c]{\textbf{Networks \& } \\ \textbf{Distributed} \\ \textbf{Systems} }} & \textbullet\ an implementation and evaluation ... mobile ad-hoc networks & \textbullet\  a pattern-based ... for a multi-core system development\textbf{(x)}\\
      & \textbullet\ trustworthiness among peer ... distributed agreement protocol & \textbullet\  model checking prioritized timed systems\textbf{(x)}\\
      & \textbullet\ reduction of processing ... synchronize multmedia replicas &  \textbullet\  learning-based adaptation to ... reconfigurable network-on-chip \textbf{(x)} \\
      & \textbullet\ quorum-based replication of multimedia ... distributed systems & \textbullet\  design issues in a performance ... embedded multi-core systems \textbf{(x)} \\
      & \textbullet\ dynamic clusters of servers to reduce total power consumption & \textbullet\  the architecture of parallelized cloud-based ... testing system\\
      & \textbullet\  experimentation of group communication protocols& \textbullet\ counterexample-guided assume-guarantee synthesis learning \textbf{(x)} \\
      & \textbullet\  a dynamic energy-aware server selection algorithm & \textbullet\ mechanism design ... allocation in non-cooperative cloud systems\\
      & \textbullet\  trustworthiness ... in peer-to-peer(p2p) overlay networks& \textbullet\ a qos-aware uplink scheduling paradigm for lte networks\\
      & \textbullet\ trustworthiness-based group communication protocols & \textbullet\ a cluster-based trust-aware routing protocol ... networks\\
      & \textbullet\ quorums for replication ... objects in p2p overlay networks & \textbullet\ a robust topology control solution for the sink placement problem in wsns\\
      & \textbullet\ performance of optimized link ... networks cloud computing & \textbullet\ profit and penalty aware (pp-aware) ... variable task execution time \textbf{(x)}\\
      & \textbullet\ error model guided joint ... optimization for flash memory \textbf{(x)} & \textbullet\ context reasoning using extended ... pervasive computing environments \textbf{(x)}\\
      & \textbullet\ cooperating virtual memory ... for flash-based storage systems \textbf{(x)} & \textbullet\ skimpystash: ram space skimpy ... flash-based storage \textbf{(x)}\\
      & \textbullet\ write mode aware loop tiling ... low power volatile pcm \textbf{(x)} & \textbullet\ flashstore: high throughput persistent key-value store \textbf{(x)}\\
      & \textbullet\ scheduling to optimize cache ... non-volatile main memories \textbf{(x)} & \textbullet\ implementation of a virtual switch monitor system using openflow on cloud\\ 
      & \textbullet\ retention trimming for wear ... flash memory storage systems \textbf{(x)} & \textbullet\ traffic load analysis ... on ieee 802.15.4/zigbee sensor network\\
      & \textbullet\ trustworthiness in p2p: ... systems for jxta-overlay platform & \textbullet\ cloud computing in taiwan\\
      & \textbullet\ a place-aware stereotypical trust supporting scheme & \textbullet\ implementation of load balancing method for cloud service with open flow\\
      & \textbullet\ price-based congestion ... multi-radio wireless mesh networks & \textbullet\ acsp: a novel security protocol against counting attack for uhf rfid systems\\
      & \textbullet\ hiam: hidden ... multi-channel multi-radio mesh networks & \textbullet\ implementation of a dynamic adjustment ... transfer in co-allocation data grids\\
    \bottomrule
\end{tabular}
}
\end{table*}

\end{document}